\documentclass[11pt]{article}

\usepackage{acl}

\usepackage{times}
\usepackage{latexsym}

\usepackage[T1]{fontenc}

\usepackage[utf8]{inputenc}

\usepackage{microtype}

\usepackage{inconsolata}

\usepackage{graphicx}

\usepackage{booktabs}
\usepackage{multirow}
\usepackage{amsmath}
\usepackage{xcolor}
\usepackage{caption}
\usepackage{amssymb}
\usepackage{subfigure}
\usepackage{subcaption}
\usepackage{color}
\usepackage[table]{xcolor}

\usepackage{hyperref}
\usepackage{url}

\definecolor{skyblue}{HTML}{D4F6FF}
\definecolor{lightred}{HTML}{FFE3E3}
\definecolor{lightbluegreen}{HTML}{D1F8EF}
\definecolor{lightpurple}{HTML}{EBEAFF}
\def\method{LongMab}
\usepackage{arydshln}

\title{Chunks as Arms: Multi-Armed Bandit-Guided Sampling for Long-Context LLM Preference Optimization}

\author{Shaohua Duan$^{1}$\thanks{ \ \ indicates equal contribution.}, Pengcheng Huang$^{1}$\footnotemark[1], Xinze Li$^{1}$\footnotemark[1], Zhenghao Liu$^{1}$\thanks{ \ \ indicates corresponding author.},\\ \textbf{Xiaoyuan Yi$^{2}$, Yukun Yan$^{3}$, Shuo Wang$^{3}$, Yu Gu$^{1}$, Ge Yu$^{1}$, Maosong Sun$^{3}$} \\ 
$^1$School of Computer Science and Engineering, Northeastern University, China \\
$^2$Microsoft Research Asia, China \\
$^3$Department of Computer Science and Technology, Institute for AI, Tsinghua University, China\\
}

\begin{document}
\maketitle

\begin{abstract}
Long-context modeling is critical for a wide range of real-world tasks, including long-context question answering, summarization, and complex reasoning tasks. Recent studies have explored fine-tuning Large Language Models (LLMs) with synthetic data to enhance their long-context capabilities. However, the effectiveness of such approaches is often limited by the low diversity and factual inconsistencies in the generated data. To address these challenges, we propose \method{}, a novel framework that leverages a Multi-Armed Bandit (MAB) rollout strategy to identify the most informative chunks from the given long context for sampling high-quality and diverse responses and constructing preference data pairs for Direct Preference Optimization (DPO) training. Specifically, we treat context chunks as arms of MAB, select chunks based on their expected reward scores to input into LLMs to generate responses, and iteratively update these scores based on reward feedback. Both exploration and exploitation during the rollout process enable the LLM to focus on the most relevant context segments, thereby generating and collecting high-quality and diverse responses. Experimental results on both Llama and Qwen show the effectiveness of \method{} by achieving more than a 4\% improvement on long-context reasoning benchmarks. All data and code will be released on \url{https://github.com/NEUIR/LongMab-PO}.
\end{abstract}

\section{Introduction}
\begin{figure}[!t] 
\centering
    \includegraphics[width=0.48\textwidth]{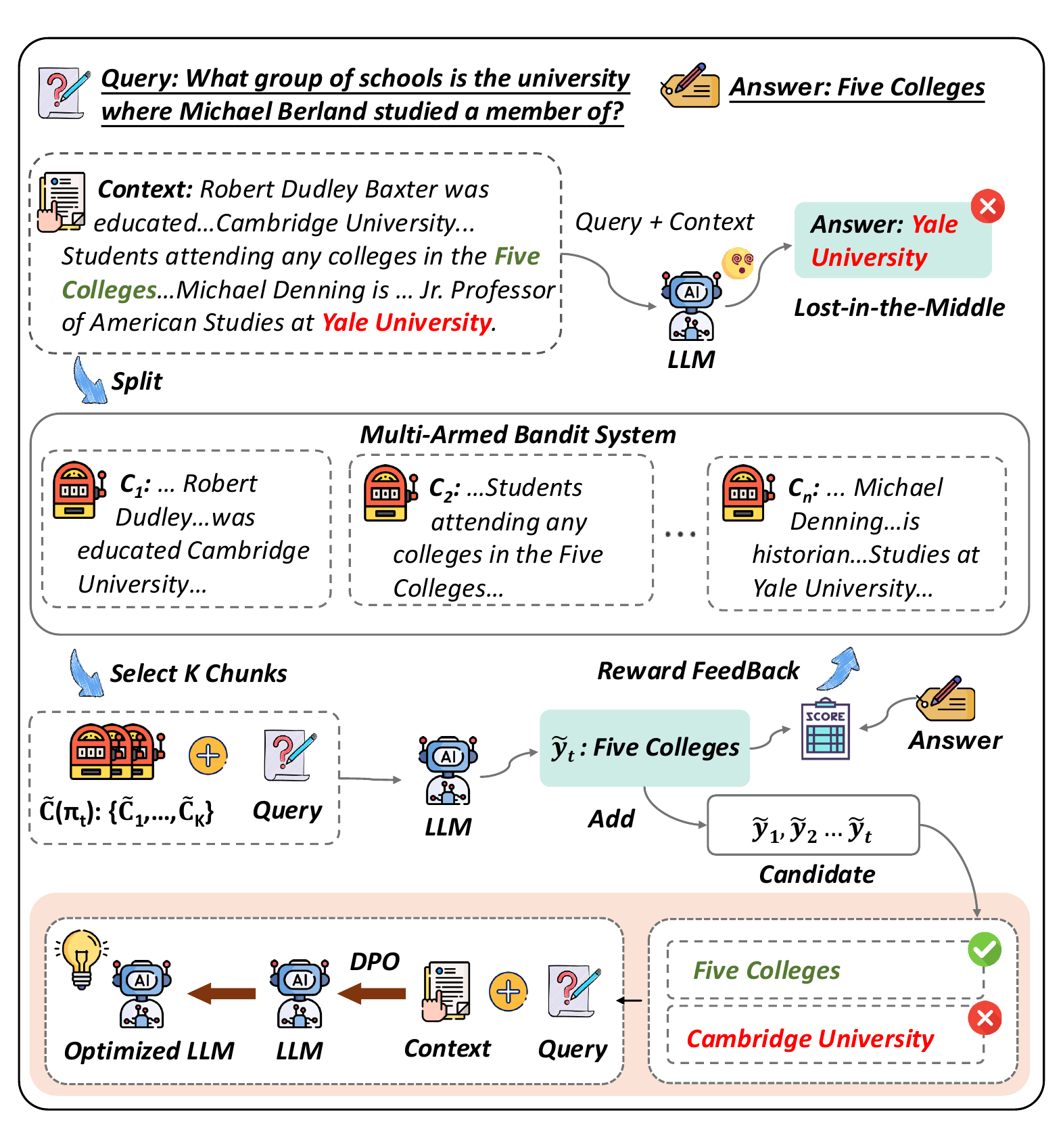}
    \caption{Illustration of the \method{} framework. \method{} conducts rollout to progressively identify informative chunk combinations and improve response quality. The collected responses are converted into preference pairs, which provide effective supervision for enhancing the model’s ability to perceive and utilize long-context information.} 
    \label{fig:LongMab_intro}
\end{figure}

Recent advancements in Large Language Models (LLMs), particularly the expansion of their context window~\citep{dubey2024llama,qwen2,liu2025autoencoding}, have enabled their application in a variety of long-context tasks~\citep{yang2025longfaith,wang2022squality,zhao2024position,liu2024forgetting}. Despite these developments, LLMs continue to suffer from the ``lost-in-the-middle'' problem~\citep{liu2023lost,he2024never}, where LLMs tend to overemphasize the beginning and end of a long input while neglecting critical information in the middle. To address this challenge, recent studies have focused on constructing high-quality Supervised Fine-Tuning (SFT) datasets specifically designed to extend the context window to long-context scenarios~\citep{chen2023longlora,bai2024longalign,chen2024essential}. Although these methods achieve certain improvements, they often induce overfitting to training-specific signals~\citep{lirag}, which in turn leads to catastrophic forgetting of general capabilities~\citep{luo2023empirical}.



An alternative line of research investigates Direct Preference Optimization (DPO)~\citep{rafailov2023direct,sun2025solopo}, which enhances the long-context understanding capability of LLMs through preference-pair training. To this end, some works~\cite{li2024large,zhang2024longreward} employ full-context sampling strategies to construct preference pairs for DPO training. However, these methods struggle to guarantee response quality, as LLMs remain highly susceptible to noise introduced by the lengthy context~\citep{shi2023large, xu2025does}.
To improve the quality of preference pair construction, recent studies~\citep{tang2024logo} introduce chunk-aware sampling strategies. These methods divide the input context into multiple chunks and construct preference pairs based on various chunk combinations: relevant chunks are used to generate positive responses, while other chunks containing noise are treated as disturbances to encourage more diverse outputs. Despite these advantages, such approaches often overlook the supportive relationships among different chunks, which limits the ability of LLMs to comprehensively identify and integrate informative content across the entire context.

In this paper, we propose \textbf{M}ulti-\textbf{a}rmed \textbf{b}andit-guided sampling for \textbf{Long}-context LLM Preference Optimization (\method{}). As illustrated in Figure~\ref{fig:LongMab_intro}, the core idea is to treat divided context chunks as the arms of a Multi-Armed Bandit (MAB) and adopt a chunk-aware sampling strategy to encourage LLMs to construct higher-quality preference pairs for DPO training. During bandit rollouts, \method{}  applies the Upper Confidence Bound (UCB) algorithm to select a subset of chunks according to their estimated rewards, thereby balancing exploration and exploitation. At each step, the selected chunks are fed into the LLM to produce a response, which is then evaluated to obtain a reward. This reward updates the expected values of the corresponding chunks, guiding subsequent selection. Through this iterative process, \method{} progressively prioritizes more informative chunk combinations, ultimately yielding diverse and high-quality responses.

The experimental results demonstrate the effectiveness of \method{}, which consistently outperforms existing SFT- and DPO-based baselines across multiple long-context tasks, achieving average gains exceeding 4\%. Further analyses show that the UCB-guided sampling strategy effectively balances exploration and exploitation, enabling the model to explore diverse chunk combinations while progressively refining the selection toward evidence-rich and complementary information. This balance ultimately leads to higher quality and greater diversity in the sampled responses, allowing \method{} to construct effective and more informative preference pairs that significantly facilitate the DPO training process.

\section{Related Work}
Numerous studies have sought to improve how Large Language Models (LLMs) utilize long contexts~\citep{hsieh2024ruler, levy2024same, wang2025lost}. Within this area, data-centric methods~\citep{chen2023longlora, zhang2024longreward, chen2025longpo}, which focus on training better models through superior data engineering—have become the mainstream approach due to their excellent performance. These strategies generally fall into two primary categories, which mainly differ in how they construct and leverage training signals.

The first of these categories, Supervised Fine-Tuning (SFT), has been a foundational approach in this domain~\citep{chen2023longlora, xiong2023effective, an2024make, xu2024chatqa, li2024making}. Early efforts relied on human annotations to create long-context question-answer pairs for training~\citep{chen2023longlora, xu2024chatqa}. However, the significant cost and scalability challenges of this manual process prompted a shift towards automated data synthesis. This is now commonly achieved by leveraging powerful LLMs to generate QA pairs from long documents, often via the self-instruct technique~\citep{wang2022self, bai2024longalign}. To further refine data quality, the most recent works have augmented this pipeline by employing LLMs as judges to filter synthetic samples, thus ensuring a more reliable training set~\citep{chen2024essential, zhu2025chain}.

As a counterpart to SFT, the other major approach leverages preference-based signals, with recent work focusing on Reinforcement Learning (RL) and particularly Direct Preference Optimization (DPO)~\citep{rafailov2023direct, yao2025expandr, li2024large}. A primary challenge in this area is sourcing high-quality preference pairs from long contexts~\citep{zhang2024longreward}. One line of work addresses this by developing sophisticated reward models, using techniques like LLM-as-a-Judge or self-consistency to generate reliable preference labels~\citep{zhang2024longreward, li2024large}. While these methods are effective, they can be hampered by contextual noise in long documents, which leads to suboptimal response generation and limits training efficacy~\citep{shi2023large}. A complementary approach is heuristic sampling, in which positive responses are generated from query-relevant chunks and negative responses from irrelevant chunks or the full context~\citep{tang2024logo, chen2025longpo, sun2025solopo}. However, this relevance-based method does not explore different chunk combinations for response generation, limiting its ability to identify the most informative evidence from the full context. To address this limitation, \method{} designs a multi-armed-bandit-guided sampling strategy that dynamically estimates the utility of chunk combinations during rollouts, constructing more informative preference pairs for DPO training.





\section{Methodology}



\begin{figure*}[t] 
\centering
    \includegraphics[width=1.0\textwidth]{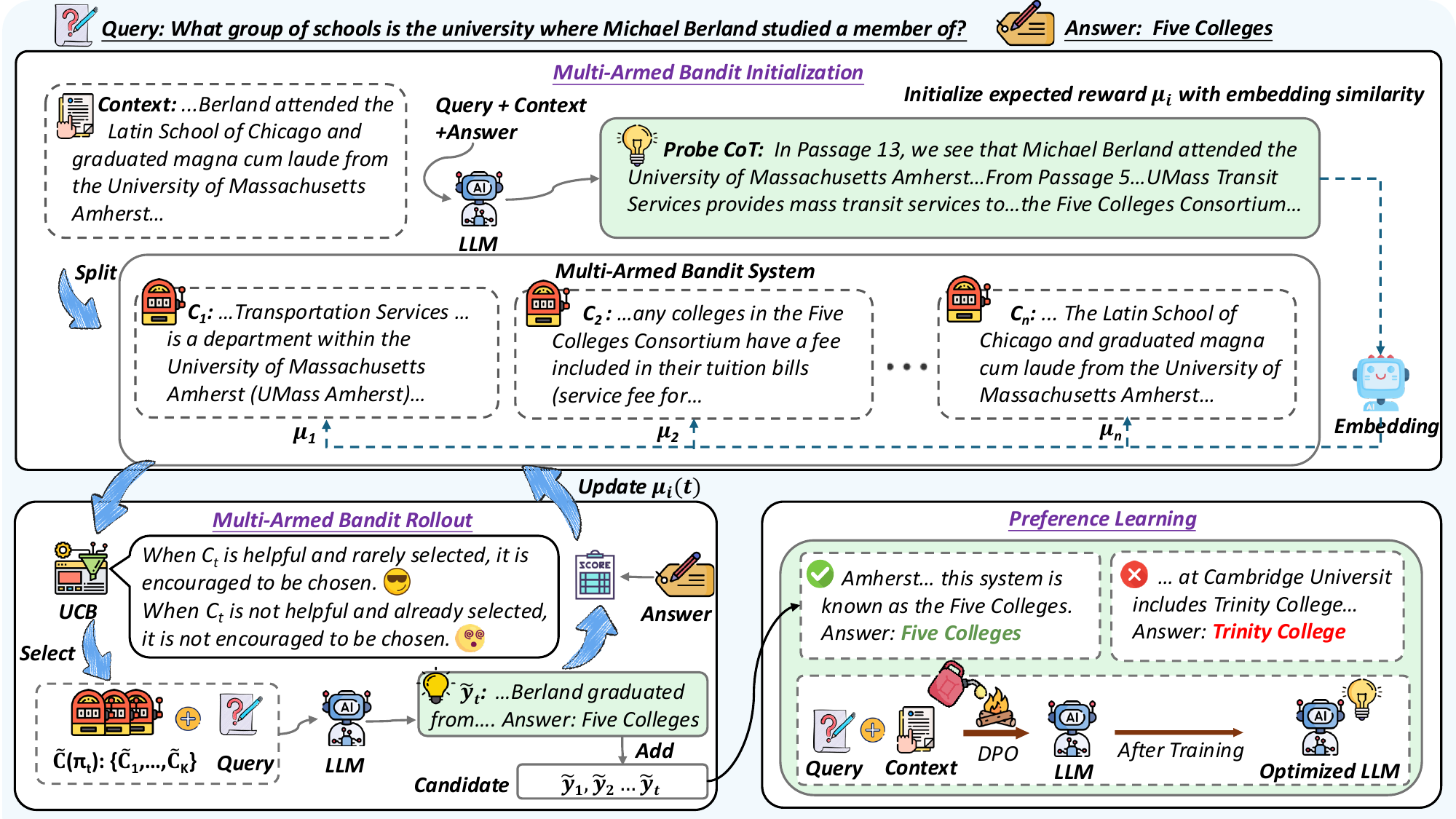}
    \caption{Illustration of multi-armed bandit-guided
sampling for long-context LLM optimization (\method{}). \method{} collects preference pairs during the rollout process for DPO training.} \label{LongMab}
\end{figure*}

As illustrated in Figure~\ref{LongMab}, this section introduces \method{}, a novel framework designed to enhance the long-context reasoning ability of language models by leveraging the Multi-Armed Bandit (MAB) paradigm. We begin by describing how a MAB rollout process is used to optimize response sampling for more effective preference learning (\S~\ref{method:training}). We then explain how the Upper Confidence Bound (UCB) algorithm is integrated to improve response quality by balancing the exploration-exploitation trade-off during the sampling process (\S~\ref{method:MABsampling}).

\subsection{Optimizing Long-Context LLMs via Multi-Armed Bandit-Guided Sampling}\label{method:training}

In long-context reasoning, a generation model $\mathcal{M}$ is tasked with generating an answer $y$ for a given question $q$ by utilizing information from a long context $C$:
\begin{equation}
\mbox{\small $y = \mathcal{M}(C,q).$}
\end{equation}

To enhance the model's long-context reasoning, we follow recent studies~\citep{zhang2024longreward} and employ DPO~\citep{rafailov2023direct} to fine-tune the model on a preference dataset $\mathcal{D}$:
\begin{equation}\label{eq:dpo}
\small
\mbox{\small $\begin{aligned}
 & \mathcal{L}_{\text{DPO}} = -\mathbb{E}_{(C,q,y^{+},y^{-}) \sim \mathcal{D}} [\log \sigma (\beta \log \\ & \frac{\mathcal{M}(y^{+} \mid C,q)}{\mathcal{M}^\text{ref}(y^{+} \mid C,q)} - \beta \log \frac{\mathcal{M}(y^{-} \mid C,q)}{\mathcal{M}^\text{ref} (y^{-} \mid C,q)})],
\end{aligned}$}
\end{equation}
where $\beta$ is a hyperparameter, $\mathcal{M}^\text{ref}$ is a fixed reference model and each instance $(C, q, y^+, y^-) \in \mathcal{D}$ contains a context-question pair $(C, q)$, along with a preferred $(y^+)$ and dispreferred $(y^-)$ response.

\textbf{Chunk-Aware Response Sampling.}
To generate high-quality preference data for more effective training, we introduce a chunk-aware response sampling framework.

Specifically, we first divide the long context $C$ into $n$ equal-length chunks, denoted as $\mathcal{C}_{\text{chunk}} = \{C_1, C_2, \dots, C_n\}$. Instead of generating a response from the entire, noisy long context $C$, our method samples from a subset of $K$ chunks, $\Tilde{\mathcal{C}}(\pi) = \{\Tilde{C}_1, \dots, \Tilde{C}_K \}$, selected by a policy $\pi$. The response is then generated as follows:
\begin{equation}
\mbox{\small $\Tilde{y} = \mathcal{M}(\Tilde{\mathcal{C}}(\pi),q).$}
\end{equation}
This sampling process is repeated for $T$ iterations. In each step, the selection policy $\pi$ evolves to explore different chunk combinations, with the goal of identifying the most informative subsets. This procedure yields a diverse set of candidate responses, $\Tilde{Y} = \{\Tilde{y}_1, \dots, \Tilde{y}_T\}$, which are then scored using the following reward function $r(\cdot)$ to construct preference pairs:
\begin{equation}
\mbox{\small $r(\Tilde{y}) = (\text{SubEM}(\Tilde{y}, y^*) + \text{F1}(\text{Ans}(\Tilde{y}),y^*))/2,$}
\end{equation}
where $y^*$ is the ground-truth answer and SubEM/F1 are reasoning quality metrics. The winning response ($y^+$) and losing response ($y^-$) are the candidates from $\Tilde{Y}$ with the highest and lowest reward scores, respectively, which completes the preference tuple $(C, q, y^+, y^-)$ for tuning. 

\textbf{Optimize Chunk Selection via Multi-Armed Bandit Rollouts.} 
To further improve the chunk-aware response sampling process, we design a chunk selection strategy $\pi$ based on MAB rollouts to adaptively explore useful chunks. 

Specifically, we model the chunk selection process as an MAB problem, where each chunk $C_i \in \mathcal{C}_{\text{chunk}}$ is treated as an individual arm. The process unfolds over $T$ rollout steps. In each step $t \in \{1, \dots, T\}$, the MAB policy $\pi_t$ selects a subset of $K$ chunks with the highest estimated expected rewards, $\Tilde{\mathcal{C}}(\pi_t)$. This subset is then used along with the question $q$ to prompt the LLM, generating a single response $\Tilde{y}_t$. After $T$ rollouts, these responses are collected to form the final candidate set, $\Tilde{Y} = \{\Tilde{y}_1, \dots, \Tilde{y}_T\}$.
\begin{equation}
\small
\mbox{\small $\Tilde{Y} = \{\Tilde{y}_1, \dots, \Tilde{y}_T\}, \quad \text{where} \quad \Tilde{y}_t = \mathcal{M}(\Tilde{\mathcal{C}}(\pi_t), q),$}
\end{equation}
where $\Tilde{y}_t$ denotes the response generated by the LLM at rollout step $t$, and $\Tilde{Y}$ are used to construct preference pairs for training. In the rollout process, the expected reward of a chunk $\Tilde{C}_i \in \Tilde{\mathcal{C}}(\pi_t)$ is determined by the quality of the response $\Tilde{y}_t$. If the selected chunks $\Tilde{C}_i$ lead the LLM to generate high-quality responses, they will receive higher expected rewards. Consequently, the MAB strategy is more likely to select more informative chunks that win a higher expected reward in the next round, thereby continuously enhancing the quality of the generated responses.

\subsection{Adaptive Chunk Prioritization through Multi-Armed Bandit Rollouts} \label{method:MABsampling}
The MAB rollout process must contend with the classic exploration–exploitation dilemma~\cite{auer2002finite}. Exploring a broader set of chunk combinations increases the chance of uncovering highly rewarding evidence but comes at a substantial computational cost. On the other hand, limited exploration may overlook informative chunks that are crucial for generating high-quality responses. Thus, it is essential to strike a careful balance between exploration (sampling under-visited chunks) and exploitation (favoring chunks with high expected rewards).

To solve this decision-making problem, we use the Upper Confidence Bound (UCB) algorithm~\citep{komiyama2024finite} to guide chunk selection at each step. Specifically, at the $t$-th rollout step, the UCB score for chunk $C_i$ is computed as:
\begin{equation}\label{eq:ucb}
\mbox{\small $\text{UCB}_t(C_i) = \mu_i (t) + \alpha \cdot \sqrt{{2 \ln{t}}/({n_i (t) + \epsilon})},$}
\end{equation}
where $\mu_i(t)$ is the expected reward of chunk $C_i$ (initialized via a probe-based strategy discussed later), $n_i(t)$ is its selection count, and $t$ is the current rollout step. The hyperparameter $\alpha$ serves to balance exploration and exploitation, while the constant $\epsilon$ prevents division by zero. After scoring all chunks in $\mathcal{C}_{\text{chunk}}$, we form the subset for the current rollout, $\Tilde{\mathcal{C}}(\pi_t)$, by selecting the top-$K$ chunks with the highest UCB scores:
\begin{equation}\label{eq:topk}
\mbox{\small $\Tilde{\mathcal{C}}(\pi_t) = \text{TopK}_{C_i \in \mathcal{C}_{\text{chunk}}} \text{UCB}_t(C_i).$}
\end{equation}
The UCB score of each chunk is updated during the rollout process based on the current expected reward of the chunk. Then, we will provide a detailed explanation of how to initialize the expected reward for each chunk and the update process of each chunk's UCB score.





\textbf{Multi-Armed Bandit Initialization via Evidence Probing.} 
To alleviate the cold-start problem in multi-armed bandit rollouts~\citep{wang2025clusterucb}, we adopt a probe-based initialization strategy~\citep{chen2025clueanchor} that assigns informed expected rewards $\mu_i(1)$ to each chunk $C_i$ prior to the selection process in rollout step $t=1$.




Specifically, we prompt the LLM $\mathcal{M}$ to generate a faithful reasoning trace, $y_{\text{Probe}}$, that explicitly identifies the evidence in the long context $C$ required to derive the ground-truth answer, $y^*$:
\begin{equation}
\mbox{\small $y_{\text{Probe}} = \mathcal{M}(\text{Instruct}_{\text{extract}}(q, \mathcal{C}, y^*)),$}
\end{equation}
where $\text{Instruct}_{\text{extract}}$ represents the instruction for extracting evidence from $\mathcal{C}$. We then score each chunk $C_i$ against the probe trace $y_{\text{Probe}}$ using cosine similarity in their embedding space:
\begin{equation}\small
s_i = \text{cos}(\text{Emb}(y_{\text{Probe}}), \text{Emb}(C_i)).
\end{equation}
These similarity scores $s_i$ then become the initial expected rewards $\mu_i(1)$. Consequently, the initial policy $\pi_1$ selects the top-$K$ chunks with the highest scores, forming the subset $\Tilde{\mathcal{C}}(\pi_1)$. This strategy provides the MAB policy with a strong semantic prior, improving early-stage efficiency and reducing blind exploration.



\textbf{UCB Score Update with MAB Rollouts.}
In each rollout step $t$ of the MAB rollout process, we need to update the corresponding UCB scores of the chunks based on their expected rewards. During the $t$-th sampling step, we prompt the LLM $\mathcal{M}$ to answer the question $q$ based on the selected chunks $\Tilde{\mathcal{C}}(\pi_t)$:
\begin{equation}
\mbox{\small $\Tilde{y}_t = \mathcal{M}(\Tilde{\mathcal{C}}(\pi_t),q),$}
\end{equation}
We then evaluate the utility of the selected chunks $\Tilde{\mathcal{C}}(\pi_t)$ by computing a reward score $r(\Tilde{y}_t)$ for the generated response $\Tilde{y}_t$. Next, we update the UCB statistics based on the reward $r(\Tilde{y}_t)$.
For each chunk $C_i \in \mathcal{C}_{\text{chunk}}$, we update its selection count at the end of rollout step $t$:
\begin{equation}
\small
\mbox{\small $n_i(t+1) =
\begin{cases}
n_i(t) + 1 & \text{if } C_i \in \Tilde{\mathcal{C}}(\pi_t), \\
n_i(t) & \text{otherwise},
\end{cases}$}
\end{equation}
and update its expected reward $\mu_i$ using an incremental average:
\begin{equation}
\small
\mbox{\small $\mu_i(t+1) =
\begin{cases}
\frac{1}{t} \left( \mu_i(t) \cdot (t - 1) + r(\Tilde{y}_t) \right) & \text{if } C_i \in \Tilde{\mathcal{C}}(\pi_t), \\
\mu_i(t) & \text{otherwise}.
\end{cases}$}
\end{equation}
The updated values $n_i(t+1)$ and $\mu_i(t+1)$ are subsequently used to compute UCB scores for the next rollout step $t+1$ using Eq.~\ref{eq:ucb}, enabling the bandit policy to continuously refine its estimation of chunk utility based on observed response quality.

\section{Experimental Methodology}
In this section, we describe the datasets, baselines, evaluation metrics, and implementation details.


\textbf{Datasets.} In our experiments, we use the MuSiQue training dataset~\citep{trivedi2022musique} to sample responses and construct preference data pairs for DPO training. To better simulate long-context scenarios, we follow the methodology of previous work~\citep{li2024large, zhu2025chain}, augmenting the context with randomly sampled Wikipedia documents to extend its length to 8k-16k tokens. For evaluation, we adopt five long-context QA tasks drawn from two widely used benchmarks: LongBench~\citep{bai2024longbench} and InfiniteBench~\citep{zhang2024bench}. Our evaluation suite includes four datasets from LongBench, including MuSiQue~\citep{trivedi2022musique}, 2WikiMultihopQA~\citep{ho2020constructing}, MultiFieldQA-En~\citep{bai2024longbench}, and NarrativeQA~\citep{kovcisky2018narrativeqa}, along with the En.QA task from InfiniteBench. Detailed statistics for these test sets are provided in Appendix~\ref{appendix:testsets}.

\textbf{Baselines.}  For a comprehensive evaluation of LongMab, we benchmark its performance against three distinct classes of baselines: (1) vanilla LLMs, (2) SFT-based methods, and (3) DPO-based methods.

Our SFT baselines include two established models, LongAlpaca~\citep{chen2023longlora} and LongAlign~\citep{bai2024longalign}, which are both fine-tuned on synthetic long-context QA data. We also introduce LongMab-SFT, an in-house baseline trained exclusively on the ``chosen'' responses from the preference dataset constructed for LongMab. For the DPO comparison, we select four state-of-the-art models: LongReward-PO~\citep{zhang2024longreward}, SeaLong-PO~\citep{li2024large}, LongFaith-PO~\citep{yang2025longfaith}, and Logo-PO~\citep{tang2024logo}. These models employ diverse strategies for generating preference pairs. For instance, LongReward-PO and SeaLong-PO rely on an LLM-as-a-Judge and self-consistency, respectively, to create reliable labels. LongFaith-PO, in contrast, leverages attributed prompts to elicit positive responses from relevant passages while generating negative ones from the full, ungrounded document. Finally, Logo-PO adopts a chunk-based approach, sampling positive responses from relevant document chunks and constructing negative ones by incrementally adding irrelevant chunks.

\begin{table*}[!t]
\centering
\small
\resizebox{\textwidth}{!}{
    \begin{tabular}{lccccccccccccccc}
    \hline
    \multirow{2}{*}{\textbf{Model}} &
    \multicolumn{2}{c}{\textbf{MuSiQue}} &
    \multicolumn{2}{c}{\textbf{2WikiMQA}} &
    \multicolumn{2}{c}{\textbf{MFQA-En}} &
    \multicolumn{2}{c}{\textbf{NarrativeQA}} &
    \multicolumn{2}{c}{\textbf{En.QA}} &
    \multicolumn{2}{c}{\textbf{Avg.}} \\
    ~ & SubEM & F1 & SubEM & F1 & SubEM & F1 & SubEM & F1 & SubEM & F1 & SubEM & F1   \\\hline
    \rowcolor{gray!10}\multicolumn{13}{c}{\textit{Llama-3.1-8B-Instruct}} \\
    \hline
    Vanilla LLM & 33.50 & 36.56 & 69.50 & 65.45 & 18.66 & 44.34 & 18.00 & 26.61 & 26.49 & 32.19 & 33.23 & 41.03  \\
    LongAlpaca & 34.00 & 38.36 & 69.50 & 63.64 & 16.66 & 40.67 & 16.50 & 26.88 & \underline{27.35} & 30.58 & 32.80 & 40.03 \\
    LongAlign & 36.50 & 37.62 & 69.00 & 59.35 & \underline{22.00} & \underline{48.49} & \underline{18.50} & 26.38 & 23.93 & 24.85 & 33.99 & 39.34 \\
    LongMab-SFT & 35.00 & 39.60 & 69.00 & 65.45 & 21.33 & 45.03 & 15.50 & 24.98 & 23.93 & 27.33 & 32.95 & 40.48 \\
    LongReward-PO & 40.50 & 41.41 & 68.50 & 63.23 & 20.00 & 44.70 & 16.50 &	25.92 &	26.21 &	24.20 &	34.34 & 39.89 \\
    SeaLong-PO & 37.50 & 40.52 & 68.00 & 67.07 & 19.33 & 44.12 & 16.00 & 26.45 & 27.06 & 32.06 & 33.58 & 42.04 \\
    Logo-PO & 43.50 & 46.58 & 63.00 & 61.74 & 19.00 & 44.39 & \underline{18.50} & \underline{28.05} & 26.67 & \underline{32.69} & 34.13 & 42.69 \\
    LongFaith-PO & \underline{44.00} & \underline{49.23} & \underline{75.50} & \textbf{72.80} & 20.66 & 48.10 & 9.50 & 23.26 & 24.30 & 22.35 & \underline{34.79} & \underline{43.15} \\
    LongMab & \textbf{50.00} & \textbf{52.15} & \textbf{76.00} & \underline{68.60} & \textbf{26.00} & \textbf{51.26} & \textbf{20.00} & \textbf{28.61} & \textbf{32.76} & \textbf{36.55} & \textbf{40.95} & \textbf{47.43} \\
    \hline
    \rowcolor{gray!10}\multicolumn{13}{c}{\textit{Qwen-2.5-7B-Instruct}} \\\hline
    Vanilla LLM & 33.50 & 30.52 & 58.00 & 50.14 & \underline{28.00} & 45.12 & 15.00 & 18.29 & 25.64 & 22.38 & 32.03 & 33.29 \\
    LongAlpaca & 32.50 & 33.90 & 55.50 & 50.75 & 24.66 & 45.88 & \underline{17.50} & 19.72 & \underline{29.34} & 23.82 & 31.90 & 34.81 \\
    LongAlign & 28.50 & 31.09 & 52.00 & 52.02 & 23.33 & \textbf{49.76} & 15.50 & 21.98 & 25.07 & 25.61 & 28.88 & 36.09 \\
    LongMab-SFT & 36.50 & 39.32 & 59.50 & \underline{58.34} & 21.33 & 44.06 & 15.50 & \underline{22.17} & 26.29 & 28.01 & 31.82 & \underline{38.38} \\
    LongReward-PO & 37.50 & 33.37 & 62.00 & 50.75 & \textbf{29.33} & 44.89 & 15.50 & 17.67 & 28.20 & 21.77 & 34.51 & 33.69 \\
    SeaLong-PO & 43.00 & 22.09 & \underline{67.00} & 36.39 & 27.33 & 41.08 & \textbf{18.00} & 16.27 & 28.20 & 16.95 & \underline{36.71} & 26.56 \\
    Logo-PO & 41.00 & 36.48 & 63.00 & 57.20 & 24.67 & \underline{49.18} & 15.50 & 18.28 & 28.00 & \underline{28.47} & 34.43 & 37.92 \\
    LongFaith-PO & \textbf{48.50} & \textbf{43.38} & 66.00 & 53.93 & 24.67 & 38.08 & 12.00 & 17.03 & 23.07 & 18.55 & 34.85 & 34.19 \\
    LongMab & \underline{44.00} & \underline{43.25} & \textbf{67.50} & \textbf{62.97} & 25.33 & 48.07 & \textbf{18.00} & \textbf{25.14} & \textbf{30.48} & \textbf{31.88} & \textbf{37.06} & \textbf{42.26} \\
    \hline
    \end{tabular}
}
\caption{Overall performance of Llama-3.1-8B-Instruct and Qwen-2.5-7B-Instruct on different long-context understanding tasks. The \textbf{best} and \underline{second best} results are highlighted.}
\label{tab:overall}
\end{table*}

\textbf{Evaluation Metrics.} Following previous work~\citep{li2024large,yang2025longfaith}, we employ two complementary rule-based metrics for evaluation. Specifically, the substring exact match (SubEM) score checks whether the golden answer appears as a substring in the model’s output. While widely used, it can be hacked by generating overly long responses that increase the chance of including the golden answer~\citep{yang2025longfaith}. To provide a more robust evaluation, we additionally report F1 score, which captures token-level overlap between the prediction and the golden answer by computing the harmonic mean of precision and recall.



%



\textbf{Implementation Details.} In line with previous studies~\citep{li2024large, yang2025longfaith}, we conduct experiments using Llama-3.1-8B-Instruct~\citep{dubey2024llama} and Qwen-2.5-7B-Instruct~\citep{qwen2} as backbone models. For training, we set the learning rate to $2 \times 10^{-5}$ and train each model for 2 epochs. To ensure training efficiency, we leverage Low-Rank Adaptation (LoRA)~\citep{hulora} and the LLaMA Factory framework. During the sampling phase, we configure the exploration-exploitation factor $\alpha = 1.0$ and the maximum number of rollout steps ($T$) to 30. In each step, we select $K=4$ context chunks, with each chunk comprising 1,500 tokens. More experimental details are provided in Appendix~\ref{appendix:addition_ex} and Appendix~\ref{appendix:param_analysis}.

\begin{table*}[!t]
\centering
\small
\resizebox{\textwidth}{!}{
    \begin{tabular}{lccccccccccccccc}
    \hline
    \multirow{2}{*}{\textbf{Model}} &
    \multicolumn{2}{c}{\textbf{MuSiQue}} &
    \multicolumn{2}{c}{\textbf{2WikiMQA}} &
    \multicolumn{2}{c}{\textbf{MFQA-En}} &
    \multicolumn{2}{c}{\textbf{NarrativeQA}} &
    \multicolumn{2}{c}{\textbf{En.QA}} &
    \multicolumn{2}{c}{\textbf{Avg.}} \\
    ~ & SubEM & F1 & SubEM & F1 & SubEM & F1 & SubEM & F1 & SubEM & F1 & SubEM & F1   \\\hline
    \rowcolor{gray!10}\multicolumn{13}{c}{\textit{Llama-3.1-8B-Instruct}} \\\hline
    LongMab & \textbf{50.00} & \textbf{52.15} & \textbf{76.00} & \underline{68.60} & \textbf{26.00} & \textbf{51.26} & \underline{20.00} & 28.61 & \textbf{32.76} & 36.55 & \textbf{40.95} & \textbf{47.43} \\
    LongMab (w/o Init.) & 43.50 & 49.59 & 71.50 & \textbf{68.66} & 23.33 & 46.28 & 17.00 & 29.39 & 29.71 & 33.46 & 37.01 & 45.48 \\
    LongMab (w/o UCB) & 42.00 & 47.62 & 69.00 & 65.28 & 20.67 & 47.00 & 19.50 & 27.95 & 27.71 & 33.11 & 35.78 & 44.19 \\
    Rejected w/o UCB & 43.50 & 45.16 & \underline{73.50} & 63.63 & \underline{24.67} & \underline{48.05} & 18.50 & 26.66 & 29.14 & 31.15 & \underline{37.86} & 42.93 \\
    Chosen w/o UCB & 42.50 & 48.84 & 66.00 & 66.07 & 16.67 & 44.32 & \textbf{21.00} & \textbf{32.57} & 28.57 & \underline{36.84} & 34.95 & \underline{45.73} \\
    \hdashline
    Direct-PO & \underline{46.50} & \underline{50.82} & 62.00 & 58.74 & 21.33 & 47.12 & 19.50 & \underline{31.45} & \underline{30.19} & \textbf{37.31} & 35.90 & 45.09 \\
    \hline
    \rowcolor{gray!10}\multicolumn{13}{c}{\textit{Qwen-2.5-7B-Instruct}} \\\hline
    LongMab & \underline{44.00} & 43.25 & \textbf{67.50} & \textbf{62.97} & \textbf{25.33} & \textbf{48.07} & 18.00 & \underline{25.14} & \underline{30.48} & \textbf{31.88} & \textbf{37.06} & \textbf{42.26} \\
    LongMab (w/o Init.) & \textbf{44.50} & \underline{44.06} & \underline{65.00} & 56.57 & \underline{25.00} & \underline{47.88} & \underline{18.50} & 21.53 & \textbf{31.05} & 29.70 & \underline{36.81} & 39.95 \\ 
    LongMab (w/o UCB)  & 41.00 & 41.29 & 62.50 & 54.53 & \textbf{25.33} & 47.01 & 16.00 & 21.69 & 28.86 & 29.30 & 34.74 & 38.76 \\
    Rejected w/o UCB & 42.50 & \textbf{45.33} & 60.00 & 56.50 & 24.00 & 45.93 & \underline{18.50} & 24.10 & 27.43 & 29.47 & 34.49 & \underline{40.27} \\
    Chosen w/o UCB & 41.50 & 43.39 & 62.50 & \underline{57.05} & 21.33 & 43.45 & 17.50 & 21.65 & 29.43 & 30.12 & 34.45 & 39.13 \\
    \hdashline
    Direct-PO & 38.50 & 41.17 & 59.00 & 55.46 & 21.33 & 45.24 & \textbf{19.00} & \textbf{25.21} & 27.43 & \underline{30.25} & 33.05 & 39.47 \\ 
    \hline
    \end{tabular}
}
\caption{Ablation study results on various long-context understanding tasks. The results demonstrate the effectivess of UCB-guided sampling strategy and probe-based probability initialization.}
\label{tab:ablation}
\end{table*}

\begin{table}[t]
\centering
\small
\resizebox{\columnwidth}{!}{
    \begin{tabular}{l|cccccccccccccc}
    \hline
    \multirow{2}{*}{\textbf{$K$}} &
    \multicolumn{2}{c}{\textbf{MuSiQue}} &
    \multicolumn{2}{c}{\textbf{MFQA-En}} &
    \multicolumn{2}{c}{\textbf{Avg.}} \\
    ~ & SubEM & F1 & SubEM & F1 & SubEM & F1 \\
    \hline
    1 & 44.00 & 46.81 & 21.33 & 49.09 & 38.07 & 44.85 \\
    2 & \underline{48.00} & 47.55 & \textbf{26.00} & \underline{50.29} & \underline{39.53} & 44.81 \\
    3 & 47.50 & 48.05 & \underline{23.33} & 49.00 & 38.71 & 44.89 \\
    4 & \textbf{50.00} & \textbf{52.15} & \textbf{26.00} & \textbf{51.26} & \textbf{40.95} & \textbf{47.43} \\
    5 & 47.00 & \underline{48.65} & 22.66 & 49.90 & 38.30 & \underline{45.40} \\
    \hline
    \end{tabular}
}
\caption{Impact of the number of selected chunks ($K$) in \method{}. We present results for two representative datasets and the average (Avg.) performance across all five datasets. Comprehensive results for all datasets are provided in Appendix~\ref{appendix:param_analysis}.}
\vspace{-1em}
\label{tab:table}
\end{table}

\section{Experiment Results}

In this section, we first present the overall performance of \method{}, followed by ablation studies to evaluate the contribution of its components. Subsequently, we analyze the selected chunks and generated responses by \method{}.

\subsection{Overall Performance}
Table~\ref{tab:overall} presents the overall performance of \method{} against baseline methods across various long-context understanding tasks.


Our experimental results demonstrate the effectiveness of \method{} in enhancing the long-context understanding capabilities of LLMs by consistently outperforms all baselines. Notably, although trained exclusively on the MuSiQue dataset, \method{} exhibits strong generalization ability, maintaining consistent gains across multiple out-of-domain tasks. Among different optimization strategies, SFT-based methods, such as LongAlpaca~\citep{chen2023longlora} and LongAlign~\citep{bai2024longalign}, show identical performance compared with the Vanilla LLM. This suggests that simply overfitting to ground-truth answers is insufficient for enabling LLMs to effectively identify and extract salient information from long contexts. In contrast, DPO-based methods generally outperform SFT-based ones in improving long-context understanding. Furthermore, \method{} achieves an improvement of over 4\% compared to other DPO-based approaches, which can be attributed to its Multi-Armed Bandit–guided mechanism. This mechanism facilitates the generation of higher-quality and more informative preference data, thereby enhancing the capability of LLMs to fully leverage information within extended contexts during the DPO training stage.

\subsection{Ablation Studies}
In this section, we first evaluate the role of different components involved in chunk sampling of our \method{}, and then investigate the impact of the number of selected chunks ($K$) during sampling.

As shown in Table~\ref{tab:ablation}, we analyze the contribution of each component in \method{}. First, \method{} (w/o Init.) removes the probability initialization step by setting all initial chunk rewards to zero. To further examine the impact of the UCB score in chunk sampling, we design three variants: \method{} (w/o UCB), \method{} (Rejected w/o UCB), and \method{} (Chosen w/o UCB). In \method{} (w/o UCB), chunks are sampled randomly to construct preference pairs for DPO training, without considering the UCB score. Similarly, \method{} (Rejected w/o UCB) and \method{} (Chosen w/o UCB) randomly select chunks to obtain the rejected and chosen responses, respectively. Finally, we also include Direct-PO, which samples responses from the complete context instead of from chunk combinations.

Compared with \method{} (w/o Init.), the full \method{} achieves over a 2\% improvement in F1 score, demonstrating the effectiveness of the evidence probe in providing a more informative prior probability to initialize the expected reward. We then explore the role of the UCB score in chunk sampling. The \method{} (w/o UCB) model exhibits performance comparable to the Direct-PO baseline, indicating that merely splitting long contexts into chunks does not enhance DPO training. Moreover, removing UCB-based sampling when selecting chosen or rejected responses leads to performance degradation, confirming the importance of UCB in guiding the selection of informative chunks for preference pair construction. Notably, \method{} (Rejected w/o UCB) consistently outperforms \method{} (w/o UCB), further illustrating that the UCB score facilitates the selection of higher-quality chosen responses, which benefits DPO optimization.

Next, we analyze the model's sensitivity to the number of selected chunks ($K$), a key hyperparameter in our MAB-guided sampling framework. We vary $K$ from 1 to 5 and evaluate the model's performance under each setting. As shown in Table~\ref{tab:table}, the results reveal a clear trend: performance improves as $K$ increases, peaking at $K=4$. This suggests that selecting too few chunks provides insufficient evidence for generating high-quality responses. However, performance slightly declines when $K=5$, indicating that $K=4$ strikes an optimal balance between evidence coverage and noise control. Selecting too many chunks tends to introduce irrelevant or noisy information, thereby reducing the effectiveness in constructing preference pairs for DPO training.


\subsection{Mechanism Behind \method{} in Enabling Long-Context LLMs}
This section provides a series of in-depth analyses to uncover the factors behind the superior performance of \method{} in enabling long-context LLMs. We compare two baseline variants: LongMab (w/o UCB) and LongMab (w/o Init.).


\begin{figure}[t]
  \centering
  \subfigure[Ground truth recall.]{
    \includegraphics[width=0.5\linewidth]{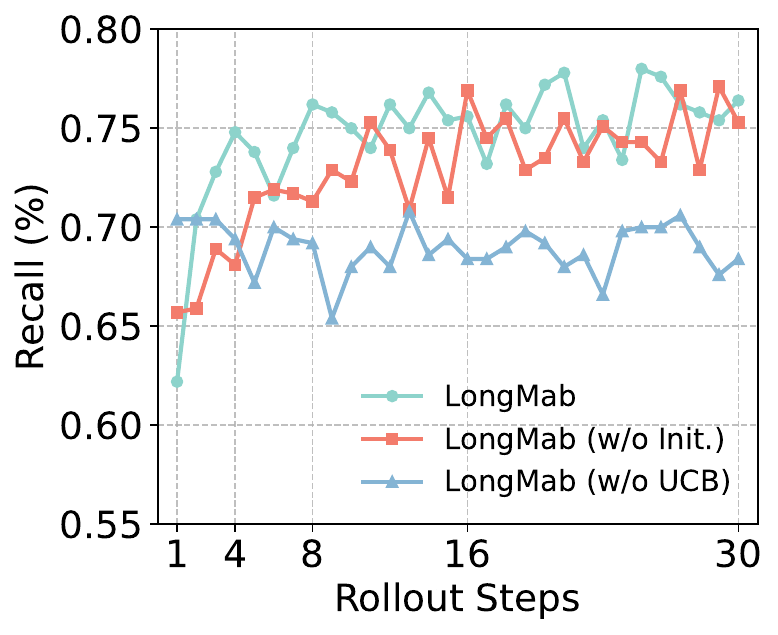}
    \label{fig:chunk_gt_recall}
  }
  \subfigure[GLM score.]{
    \includegraphics[width=0.425\linewidth]{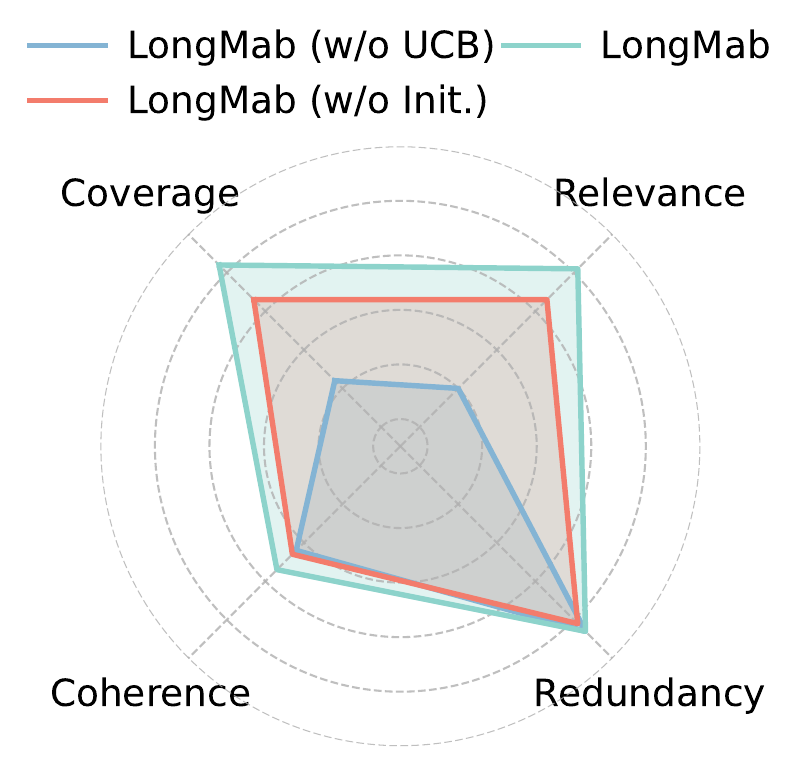}
    \label{fig:glm_eval_radar_tilted}
  }
  \subfigure[Chunk frequency.]{
    \includegraphics[width=0.43\linewidth]{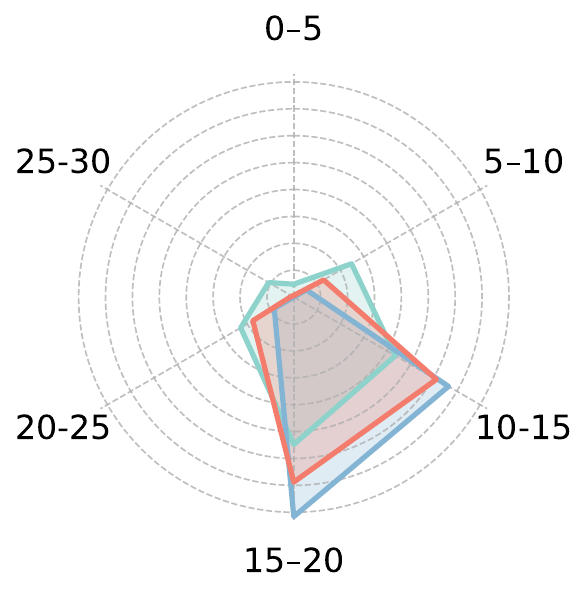}
    \label{fig:chunk_freq_radar}
  }
  \subfigure[SubEM of responses.]{
    \includegraphics[width=0.5\linewidth]{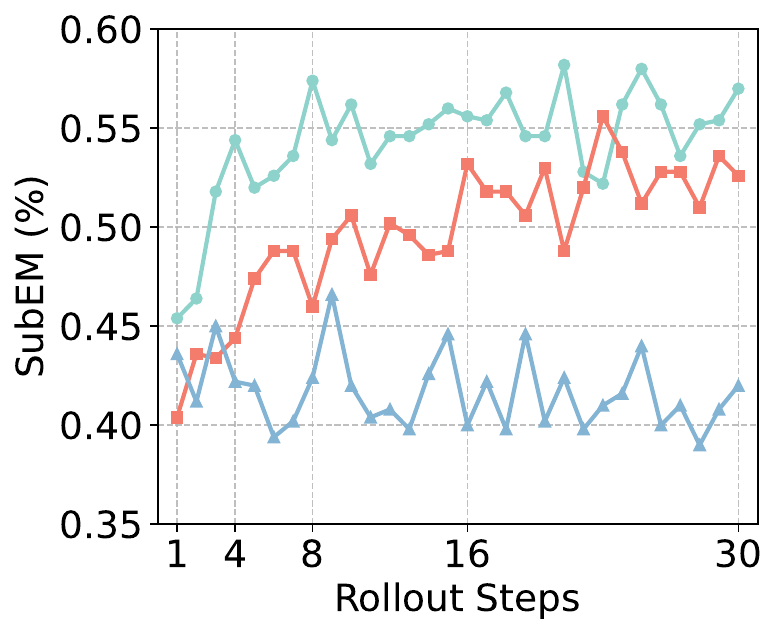}
    \label{fig:rollout_subem}
  }
  \caption{Characteristics of chunk sets obtained by different methods. Figure~\ref{fig:chunk_gt_recall} shows the ground-truth recall during the rollout process. Figure~\ref{fig:glm_eval_radar_tilted} reports GLM-4-Plus scores for the chunk sets. Figure~\ref{fig:chunk_freq_radar} illustrates the distribution of chunk selection frequencies, grouping chunks by their selection counts across all rollouts. Finally, Figure~\ref{fig:rollout_subem} presents the trend of SubEM scores over the rollout process.}
  \label{fig:chunk_analysis}
\end{figure}

\begin{figure}[t]
  \centering
  \subfigure[Average chosen SubEM.]{
    \includegraphics[width=0.46\linewidth]{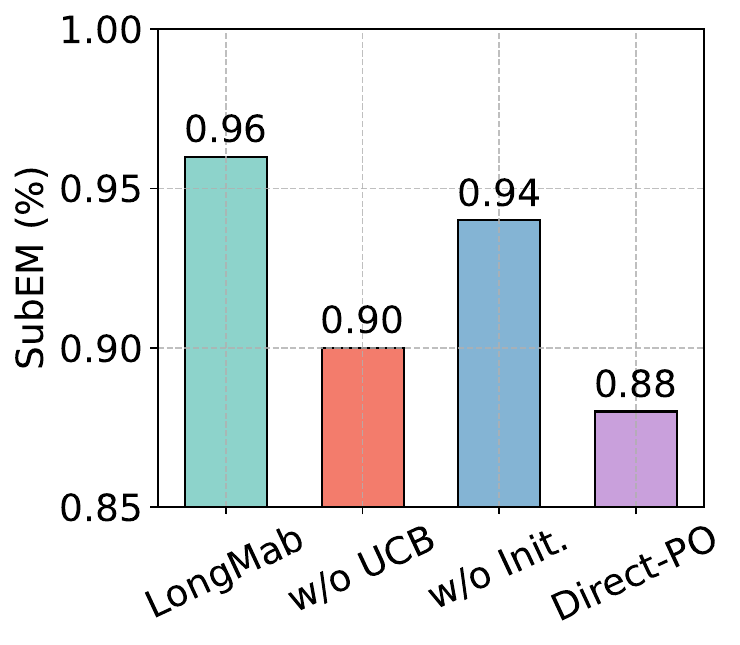}
    \label{fig:chosen_reward}
  }
  \hfill
  \subfigure[Variance of responses.]{
    \includegraphics[width=0.46\linewidth]{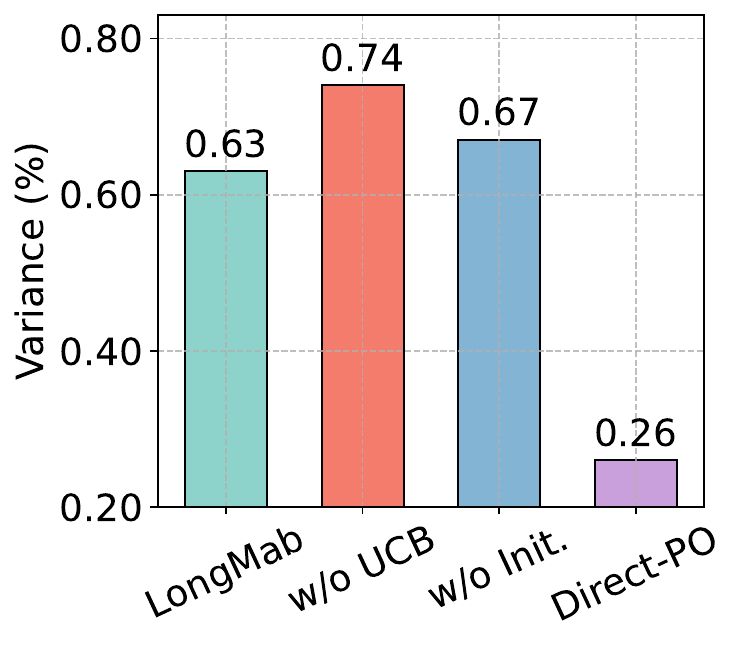}
    \label{fig:response_var}
  }
  \caption{Analysis of responses generated by different methods. Figure~\ref{fig:chosen_reward} shows the average SubEM score of the selected responses, while Figure~\ref{fig:response_var} depicts the variance of pairwise similarities among all responses in the sampled response set.}
  \label{fig:response_quality}
\end{figure}
\textbf{Characteristics of \method{}-Selected Chunks.}
As illustrated in Figure~\ref{fig:chunk_analysis}, we conduct experiments to investigate the effectiveness of the UCB-guided sampling strategy for chunk selection.

We first evaluate the quality of the selected chunks. To this end, we monitor the ground-truth recall throughout the rollout process. As shown in Figure~\ref{fig:chunk_gt_recall}, the recall of \method{} steadily increases and then stabilizes, whereas the recall of the baseline LongMab (w/o UCB) remains flat. This indicates that the UCB mechanism effectively guides the selection process toward more informative content. To further assess quality, we employ GLM-4-Plus~\citep{du2021glm} as a judge to rate the final chunk sets along four dimensions: coverage, relevance, redundancy, and coherence. As shown in Figure~\ref{fig:glm_eval_radar_tilted}, \method{} consistently outperforms all baselines across all dimensions. Notably, \method{} achieves higher scores in both coverage and relevance, underscoring its capability to form informative and well-structured chunk combinations.

Next, we analyze the selection mechanism to understand how UCB enhances chunk quality. To verify this, we examine the distribution of chunk selection frequencies shown in Figure~\ref{fig:chunk_freq_radar}. The baseline model exhibits a narrow and uniform distribution, suggesting indiscriminate sampling. In contrast, \method{} produces a broader and more polarized distribution, indicating that it simultaneously explores a diverse range of chunks while prioritizing high-value ones and suppressing less relevant options. The effectiveness of \method{} in chunk selection is further validated by the steadily increasing SubEM scores during rollout (Figure~\ref{fig:rollout_subem}), confirming that its improved selection directly lead to higher-quality responses.

\textbf{Preference Pairs Analysis.}
We next examine whether this advantage translates into constructing more effective preference data. Specifically, we aim to assess both the quality of the chosen responses and the diversity of all generated responses. As shown in Figure~\ref{fig:response_quality}, we compare \method{} with three baselines: \method{} (w/o UCB), \method{} (w/o Init.), and Direct-PO.

We first analyze the quality of the chosen responses. Figure~\ref{fig:chosen_reward} reports the SubEM scores of the selected responses from different models. The results show that \method{} achieves over a 6\% improvement compared to both \method{} (w/o UCB) and Direct-PO, demonstrating its effectiveness in generating higher-quality preferred responses.
Next, we evaluate the diversity of all sampled responses, as shown in Figure~\ref{fig:response_var}. We use the MiniCPM-Embedding model~\cite{huminicpm} to encode all generated responses and compute the average variance of pairwise similarities. \method{} yields substantially higher variance scores than Direct-PO, while maintaining a comparable level of diversity to the random sampling variant \method{} (w/o UCB). These results suggest that \method{} not only improves response quality but also preserves diversity, enabling the construction of more informative preference pairs for DPO training.

\section{Conclusion}
This paper introduces \method{}, an innovative framework that applies a Multi-Armed Bandit (MAB) approach to optimize long-context language models. Specifically, \method{} leverages an iterative MAB rollout process to identify and select optimal combinations of context chunks, enabling the sampling of higher-quality and more diverse responses. These responses are then used to construct preference data for DPO training. As a result, the aligned model exhibits a superior ability to pinpoint key information within long contexts, leading to more accurate reasoning. Our experiments show that \method{} substantially outperforms strong baseline methods across numerous long-context understanding tasks.

\section*{Limitation}
While \method{} demonstrates substantial improvements in long-context reasoning and alignment tasks, several limitations remain. First, although the UCB-guided sampling strategy effectively balances exploration and exploitation during chunk selection, its computational efficiency in extremely long contexts warrants additional optimization, particularly when a large number of rollout steps is used. Next, our experiments are conducted primarily on models around the 7B and 14B parameter scale, and the generalization of \method{} to larger LLMs remains to be verified. Finally, due to computational constraints, the synthesized instruction data used in our experiments are limited to lengths of 8K–16K tokens, and exploring the effectiveness of \method{} on even longer synthetic contexts remains an open question for future work.


\bibliography{custom}

\clearpage
\appendix
\section{Appendix}

\subsection{License}
This section summarizes the licenses of the datasets
used in our experiments. 

All of these datasets under their respective licenses and agreements allow for academic use: MuSiQue (CC-BY-4.0 license); 2WikiMultihopQA, MultiFieldQA-En, NarrativeQA, LongBench (Apache 2.0 license); En.QA, InfiniteBench (MIT license).

\subsection{Impact of Reward Design on Multi-Armed Bandit Rollout Process}
In this section, we investigate the impact of different reward designs in the multi-armed bandit rollout process on the quality of sampled responses. Specifically, we compare two different reward calculation strategies: the full response SubEM and the answer-based SubEM strategy. The first strategy computes the SubEM score based on the complete response, while the second extracts the answer string from the response and then calculates the SubEM score. In both cases, the final reward of sampled response is computed by averaging the SubEM score with the F1 score of the extracted answer.

\begin{figure}[t]
  \centering
  \subfigure[Ground truth recall.]{
    \includegraphics[width=0.46\linewidth]{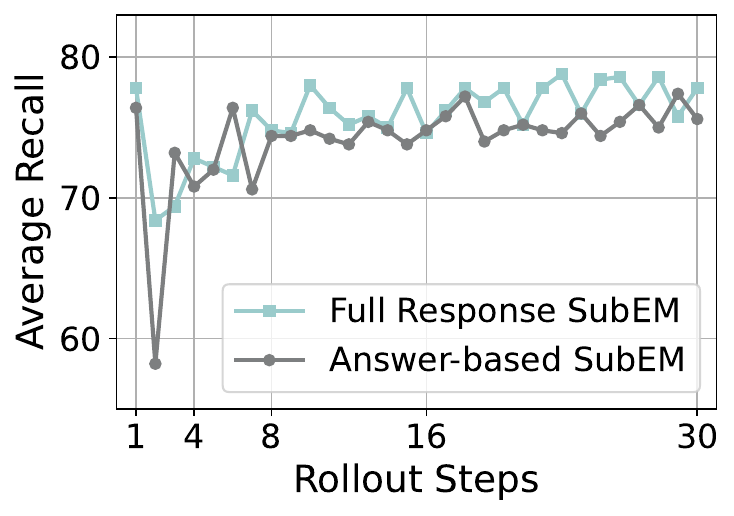}
    \label{fig:reward_recall}
  }
  \hfill
  \subfigure[Average F1 score.]{
    \includegraphics[width=0.46\linewidth]{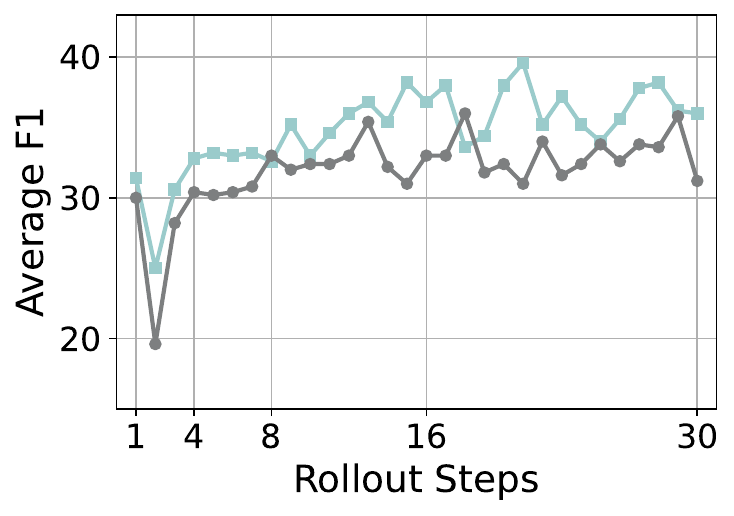}
    \label{fig:reward_f1}
  }
  \caption{Impact of different reward design to rollout process. The results show that using the full response SubEM strategy leads to consistently higher ground truth recall and F1 scores during rollout.}
  \label{fig:reward_design}
\end{figure}

As shown in Figure~\ref{fig:reward_design}, we track the ground truth recall and F1 scores of generated responses during the rollout process. The full response SubEM strategy consistently outperforms the answer-based SubEM strategy in both metrics, suggesting that overly strict reward formulations hinder exploration of diverse chunk combinations and reduce sampling quality. To further assess their impact, we train LLMs using responses sampled under each strategy and evaluate them on downstream tasks. As shown in Table~\ref{tab:reward_design}, models trained with answer-based SubEM strategy consistently underperform compared to those using full response SubEM strategy, confirming that lower sampling quality weakens subsequent DPO training.

\subsection{Additional Experimental Details}\label{appendix:addition_ex}
\begin{figure*}[t]
  \centering
  \includegraphics[width=1.0\linewidth]{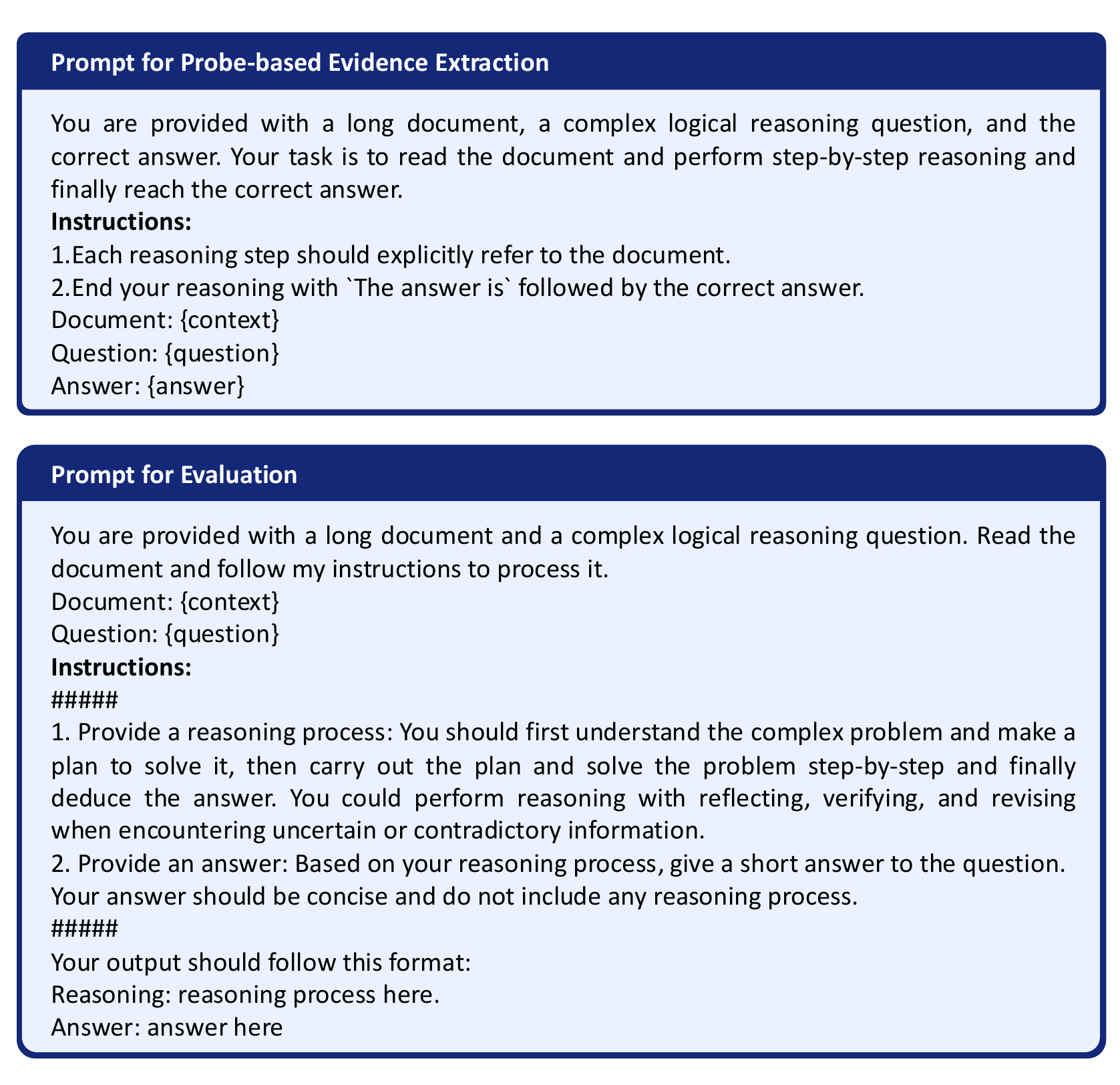}
  \caption{Prompt templates used in \method{}.}
  \label{fig:Prompt}
\end{figure*}

\begin{figure*}[t]
  \centering
  \includegraphics[width=1.0\linewidth]{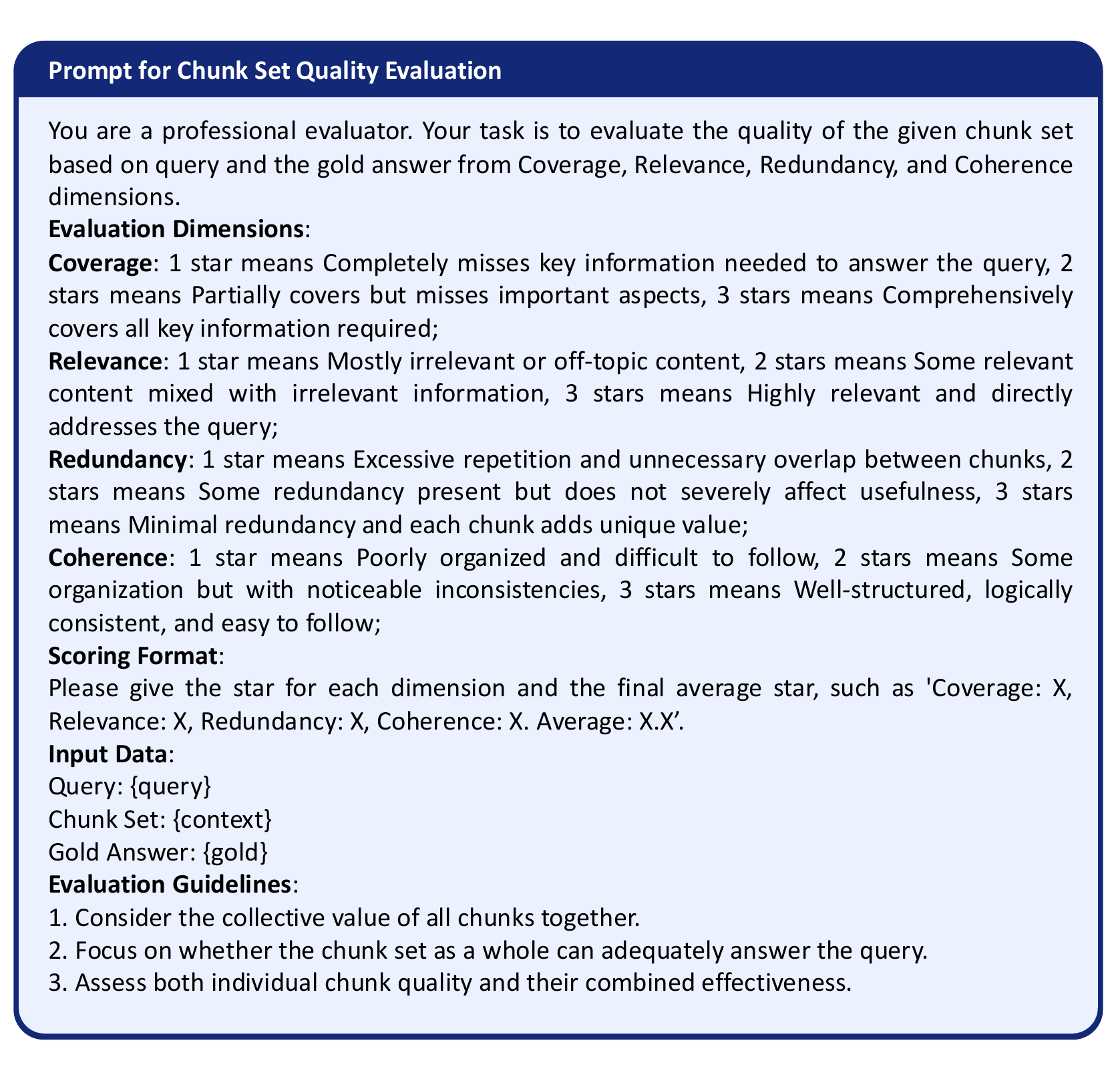}
  \caption{The prompt templates used for GLM-4-plus to evaluate the informativeness of chunk set from 4 dimensions.}
  \label{fig:prompt_chunkset_eval}
\end{figure*}

In this section, we introduce additional experimental details. In our experiments, all models use the VLLM framework during the inference stage, with the same inference settings and random seed. This ensures that all models' evaluation results in the paper are reproducible. Furthermore, during the training process, we use a unified default random seed, typically set to 42. Besides, the prompt templates used in the \method{} are shown in Figure~\ref{fig:Prompt} and Figure~\ref{fig:prompt_chunkset_eval}. All experiments were conducted on NVIDIA A100 GPUs with 80GB of memory.

\begin{table}[t]
\centering
\small
\begin{tabular}{lrrc}
\hline
                   \textbf{Dataset}  & \#Tokens & \#Samples  \\ 
                   \hline
                   MuSiQue & 15.5k & 200 \\
                    2WikiMultihopQA  & 7.1k & 200 \\ 
                    MultiFieldQA-En  & 6.9k & 150 \\
                    NarrativeQA & 29.8k & 200 \\
En.QA & 192.6k & 351 \\
\hline
\end{tabular}
\caption{The statistics of testsets from two benchmarks.}
\label{tab:benchmark_statistics}
\end{table}



\begin{table*}[t]
\centering
\small
\resizebox{\textwidth}{!}{
    \begin{tabular}{lccccccccccccccc}
    \hline
    \multirow{2}{*}{\textbf{Strategy}} &
    \multicolumn{2}{c}{\textbf{MuSiQue}} &
    \multicolumn{2}{c}{\textbf{2WikiMQA}} &
    \multicolumn{2}{c}{\textbf{MFQA-En}} &
    \multicolumn{2}{c}{\textbf{NarrativeQA}} &
    \multicolumn{2}{c}{\textbf{En.QA}} &
    \multicolumn{2}{c}{\textbf{Avg.}} \\
    ~ & SubEM & F1 & SubEM & F1 & SubEM & F1 & SubEM & F1 & SubEM & F1 & SubEM & F1   \\\hline
    \rowcolor{gray!10}\multicolumn{13}{c}{\textit{Llama-3.1-8B-Instruct}} \\
    \hline
    Full Response & \textbf{50.00} & \textbf{52.15} & \textbf{76.00} & 68.60 & \textbf{26.00} & \textbf{51.26} & \textbf{20.00} & \textbf{28.61} & \textbf{32.76} & \textbf{36.55} & \textbf{40.95} & \textbf{47.43} \\
    Answer-based  & 45.50 & 48.58 & 72.50 & \textbf{69.43} & 20.00 & 47.39 & 19.00 & 28.60 & 28.57 & 31.00 & 37.11 & 45.00 \\
    \hline
    \rowcolor{gray!10}\multicolumn{13}{c}{\textit{Qwen-2.5-7B-Instruct}} \\
    \hline
    Full Response & \textbf{44.00} & 43.25 & \textbf{67.50} & \textbf{62.97} & \textbf{25.33} & \textbf{48.07} & 18.00 & 25.14 & \textbf{30.48} & \textbf{31.88} & \textbf{37.06} & \textbf{42.26} \\
    Answer-based  & \textbf{44.00} & \textbf{44.42} & 63.00 & 56.85 & 20.66 & 46.10 & \textbf{20.50} & \textbf{25.36} & 29.62 & 30.73 & 35.56 & 40.69 \\
    \hline
    \end{tabular}
}
\caption{Performance of models trained with preference data created under different reward strategies.}
\label{tab:reward_design}
\end{table*}
\begin{table*}[t]
\centering
\small
\resizebox{\textwidth}{!}{
    \begin{tabular}{lccccccccccccccc}
    \hline
    \multirow{2}{*}{\textbf{Rollout Steps (T)}} &
    \multicolumn{2}{c}{\textbf{MuSiQue}} &
    \multicolumn{2}{c}{\textbf{2WikiMQA}} &
    \multicolumn{2}{c}{\textbf{MFQA-En}} &
    \multicolumn{2}{c}{\textbf{NarrativeQA}} &
    \multicolumn{2}{c}{\textbf{En.QA}} &
    \multicolumn{2}{c}{\textbf{Avg.}} \\
    ~ & SubEM & F1 & SubEM & F1 & SubEM & F1 & SubEM & F1 & SubEM & F1 & SubEM & F1   \\\hline
    \quad Vanilla LLM & 33.50 & 36.56 & 69.50 & 65.45 & 18.66 & 44.34 & 18.00 & 26.61 & 26.49 & 32.19 & 33.23 & 41.03 \\
    \quad\quad T=4 & 37.50 & 45.17 & 70.50 & 67.84 & \underline{24.00} & 50.56 & \underline{19.50} & \textbf{31.04} & 28.57 & \textbf{37.61} & 36.01 & \underline{46.44} \\
    \quad\quad T=8 & 45.50 & 47.75 & 75.00 & \underline{67.95} & 23.33 & 49.49 & \textbf{20.00} & 27.52 & 29.71 & 33.40 & 38.71 & 45.22 \\
    \quad\quad T=16 & \underline{48.50} & \underline{51.88} & \underline{75.50} & 66.32 & \underline{24.00} & \underline{50.84} & 19.00 & 27.66 & \underline{30.00} & 32.91 & \underline{39.40} & 45.92 \\
    \quad\quad T=30 & \textbf{50.00} & \textbf{52.15} & \textbf{76.00} & \textbf{68.60} & \textbf{26.00} & \textbf{51.26} & \textbf{20.00} & \underline{28.61} & \textbf{32.76} & \underline{36.55} & \textbf{40.95} & \textbf{47.43} \\
    \hline
    \end{tabular}
}
\caption{Scaling the rollout step of \method{}. Performance consistently increases across all testsets.}
\label{tab:rollout_step}
\end{table*}

\begin{table*}[t]
\centering
\small
\resizebox{\textwidth}{!}{
    \begin{tabular}{l|cccccccccccccc}
    \hline
    \multirow{2}{*}{\textbf{$K$}} &
    \multicolumn{2}{c}{\textbf{MuSiQue}} &
    \multicolumn{2}{c}{\textbf{2WikiMQA}} &
    \multicolumn{2}{c}{\textbf{MFQA-En}} &
    \multicolumn{2}{c}{\textbf{NarrativeQA}} &
    \multicolumn{2}{c}{\textbf{En.QA}} &
    \multicolumn{2}{c}{\textbf{Avg.}} \\
    ~ & SubEM & F1 & SubEM & F1 & SubEM & F1 & SubEM & F1 & SubEM & F1 & SubEM & F1 \\
    \hline
    1 & 44.00 & 46.81 & \textbf{76.50} & 64.81 & 21.33 & 49.09 & 18.50 & 28.40 & 30.00 & \underline{35.16} & 38.07 & 44.85 \\
    2 & \underline{48.00} & 47.55 & 73.00 & 61.73 & \textbf{26.00} & \underline{50.29} & \underline{19.50} & \textbf{29.81} & \underline{31.14} & 34.68 & \underline{39.53} & 44.81 \\
    3 & 47.50 & 48.05 & 75.50 & 67.61 & \underline{23.33} & 49.00 & 17.00 & 25.71 & 30.20 & 34.06 & 38.71 & 44.89 \\
    4 & \textbf{50.00} & \textbf{52.15} & \underline{76.00} & \textbf{68.60} & \textbf{26.00} & \textbf{51.26} & \textbf{20.00} & \underline{28.61} & \textbf{32.76} & \textbf{36.55} & \textbf{40.95} & \textbf{47.43} \\
    5 & 47.00 & \underline{48.65} & 73.50 & \underline{68.58} & 22.66 & 49.90 & 19.00 & 26.94 & 29.36 & 32.93 & 38.30 & \underline{45.40} \\
    \hline
    \end{tabular}
}
\caption{Impact of the number of selected chunks in \method{}. $K$ denotes the number of selected chunks. The \textbf{best} and \underline{second best} results are highlighted.}
\vspace{-1em}
\label{tab:top_k_appendix}
\end{table*}
\begin{table*}[t]
\centering
\small
\resizebox{\textwidth}{!}{
    \begin{tabular}{l|cccccccccccccc}
    \hline
    \multirow{2}{*}{\textbf{$\alpha$}} &
    \multicolumn{2}{c}{\textbf{MuSiQue}} &
    \multicolumn{2}{c}{\textbf{2WikiMQA}} &
    \multicolumn{2}{c}{\textbf{MFQA-En}} &
    \multicolumn{2}{c}{\textbf{NarrativeQA}} &
    \multicolumn{2}{c}{\textbf{En.QA}} &
    \multicolumn{2}{c}{\textbf{Avg.}} \\
    ~ & SubEM & F1 & SubEM & F1 & SubEM & F1 & SubEM & F1 & SubEM & F1 & SubEM & F1 \\
    \hline
    0.2 & \underline{48.50} & \underline{49.90} & \textbf{76.50} & 66.46 & 20.67 & \underline{49.65} & 17.00 & 24.90 & 27.20 & 31.27 & 37.97 & 44.44 \\
    0.5 & 47.00 & 48.49 & \textbf{76.50} & \textbf{71.38} & 23.33 & 46.72 & \textbf{21.00} & 27.17 & 28.86 & \underline{31.56} & \underline{39.34} & \underline{45.06} \\
    1.0 & \textbf{50.00} & \textbf{52.15} & \underline{76.00} & 68.60 & \textbf{26.00} & \textbf{51.26} & \underline{20.00} & \textbf{28.61} & \textbf{32.76} & \textbf{36.55} & \textbf{40.95} & \textbf{47.43} \\
    2.0 & 46.50 & 47.88 & 75.50 & \underline{68.61} & 22.00 & 46.52 & \textbf{21.00} & \underline{27.45} & \underline{30.00} & 30.54 & 39.00 & 44.20 \\
    5.0 & 40.50 & 43.71 & 73.50 & 66.19 & \underline{24.00} & 48.64 & 19.50 & 26.79 & \underline{30.00} & 29.77 & 37.50 & 43.02 \\
    \hline
    \end{tabular}
}
\caption{Sensitivity of \method{} to the exploration–exploitation factor $\alpha$. Performance first increases and then declines as $\alpha$ grows, with $\alpha=1.0$ achieving the best trade-off between exploration and exploitation. The \textbf{best} and \underline{second best} results are highlighted.}
\vspace{-1em}
\label{tab:alpha}
\end{table*}
\begin{table*}[!t]
\centering
\small
\resizebox{\textwidth}{!}{
    \begin{tabular}{lccccccccccccccc}
    \hline
    \multirow{2}{*}{\textbf{Model}} &
    \multicolumn{2}{c}{\textbf{MuSiQue}} &
    \multicolumn{2}{c}{\textbf{2WikiMQA}} &
    \multicolumn{2}{c}{\textbf{MFQA-En}} &
    \multicolumn{2}{c}{\textbf{NarrativeQA}} &
    \multicolumn{2}{c}{\textbf{En.QA}} &
    \multicolumn{2}{c}{\textbf{Avg.}} \\
    ~ & SubEM & F1 & SubEM & F1 & SubEM & F1 & SubEM & F1 & SubEM & F1 & SubEM & F1   \\\hline
    \rowcolor{gray!10}\multicolumn{13}{c}{\textit{Qwen2.5-14B-Instruct}} \\
    \hline
    Vanilla LLM & 46.50 & 39.18 & 75.50 & 63.88 & \textbf{26.67} & 47.78 & 19.50 & 22.54 & 35.04 & 24.47 & 40.64 & 39.57 \\
    LongAlpaca & 43.50 & 41.10 & 70.00 & 61.55 & \underline{25.33} & \underline{48.87} & 19.00 & 21.62 & 32.76 & 26.76 & 38.12 & 39.98 \\
    LongAlign & 37.50 & 38.33 & 68.50 & 65.26 & 20.67 & 46.25 & 17.50 & 21.52 & 32.19 & 30.65 & 35.27 & 40.40 \\
    LongMab-SFT & 45.00 & 44.00 & 73.00 & 64.01 & 23.33 & \textbf{49.96} & 20.50 & 22.67 & 31.91 & 28.42 & 38.75 & 41.81 \\
    LongReward-PO & 48.50 & 43.41 & 73.50 & 62.30 & 22.67 & 46.74 & \underline{21.00} & 20.33 & 35.61 & 27.10 & 40.26 & 39.98 \\
    SeaLong-PO & 46.50 & 37.56 & 75.00 & 57.91 & 24.67 & 46.83 & 19.00 & 21.17 & 36.47 & 25.93 & 40.33 & 37.88 \\
    Logo-PO & \textbf{51.00} & 42.04 & 77.50 & 66.52 & 26.67 & 48.19 & 20.50 & 23.41 & \textbf{36.75} & \underline{32.12} & \textbf{42.48} & 42.46 \\
    LongFaith-PO & 49.33 & \textbf{51.00} & \underline{78.00} & \underline{67.85} & 20.67 & 46.92 & 20.00 & \underline{24.37} & 34.76 & 30.93 & 40.89 & \underline{43.88} \\
    LongMab & \underline{50.00} & \underline{50.34} & \textbf{79.50} & \textbf{68.83} & 23.33 & 47.92 & \textbf{22.00} & \textbf{26.83} & \underline{35.90} & \textbf{35.29} & \underline{42.15} & \textbf{45.84} \\
    \hline
    \end{tabular}
}
\caption{Overall performance of Qwen2.5-14B-Instruct on different long-context understanding tasks. The \textbf{best} and \underline{second best} results are highlighted.}
\label{tab:qwen2_14b_overall}
\end{table*}

\begin{table*}[!t]
\centering
\small
\resizebox{\textwidth}{!}{
    \begin{tabular}{lccccccccccccccc}
    \hline
    \multirow{2}{*}{\textbf{Model}} &
    \multicolumn{2}{c}{\textbf{MuSiQue}} &
    \multicolumn{2}{c}{\textbf{2WikiMQA}} &
    \multicolumn{2}{c}{\textbf{MFQA-En}} &
    \multicolumn{2}{c}{\textbf{NarrativeQA}} &
    \multicolumn{2}{c}{\textbf{En.QA}} &
    \multicolumn{2}{c}{\textbf{Avg.}} \\
    ~ & SubEM & F1 & SubEM & F1 & SubEM & F1 & SubEM & F1 & SubEM & F1 & SubEM & F1   \\\hline
    \rowcolor{gray!10}\multicolumn{13}{c}{\textit{Qwen3-8B}} \\
    \hline
    Vanilla LLM & 41.50 & 44.24 & \underline{77.00} & 72.40 & 23.33 & 47.07 & \textbf{19.00} & \underline{26.26} & 30.29 & 31.35 & 38.22 & 44.26  \\
    LongAlpaca & 36.50 & 38.06 & 71.00 & 68.76 & 23.33 & 47.73 & 15.00 & 22.31 & 25.43 & 27.69 & 34.25 & 40.91 \\
    LongAlign & 29.00 & 28.76 & 62.00 & 54.34 & \textbf{27.33} & \textbf{50.29} & 15.00 & 22.89 & 24.86 & 26.86 & 31.64 & 36.63 \\
    LongMab-SFT & 41.50 & 45.62 & 75.50 & 73.16 & 22.67 & 47.99 & 17.00 & 25.86 & 29.43 & 32.36 & 37.22 & \underline{45.00} \\
    LongReward-PO & 40.50 & 41.94 & 76.00 & 70.08 & \underline{24.00} & 45.48 & 17.00 & 23.13 & 28.00 & 29.01 & 37.10 & 41.93 \\
    SeaLong-PO & \underline{42.50} & 44.11 & \underline{77.00} & \underline{73.38} & 22.67 & \underline{48.56} & 18.00 & 25.15 & \textbf{32.86} & \underline{33.54} & \underline{38.61} & 44.95 \\
    Logo-PO & 41.00 & 44.98 & 75.50 & 71.34 & 23.33 & 47.32 & 17.00 & 24.48 & 30.29 & 30.75 & 37.42 & 43.77 \\
    LongFaith-PO & 41.00 & \underline{46.35} & 66.00 & 66.79 & 17.33 & 38.01 & 11.50 & 18.31 & 24.29 & 28.08 & 32.02 & 39.51 \\
    LongMab & \textbf{48.00} & \textbf{51.02} & \textbf{77.50} & \textbf{74.32} & 23.33 & 47.57 & \underline{18.50} & \textbf{27.34} & \underline{31.43} & \textbf{33.84} & \textbf{39.75} & \textbf{46.82} \\
    \hline
    \end{tabular}
}
\caption{Overall performance of Qwen3-8B on different long-context understanding tasks. The \textbf{best} and \underline{second best} results are highlighted.}
\label{tab:qwen3_overall}
\end{table*}

\begin{table*}[!t]
\centering
\small
\resizebox{\textwidth}{!}{
\begin{tabular}{l c c c c c c c c c c}
\hline
\multirow{2}{*}{\textbf{Model}} &
\multicolumn{2}{c}{\textbf{16k}} &
\multicolumn{2}{c}{\textbf{32k}} &
\multicolumn{2}{c}{\textbf{64k}} &
\multicolumn{2}{c}{\textbf{128k}} &
\multicolumn{2}{c}{\textbf{Avg.}} \\
~ & SubEM & F1 & SubEM & F1 & SubEM & F1 & SubEM & F1 & SubEM & F1 \\
\hline
Vanilla LLM & 30.00 & 29.44 & 29.17 & 28.00 & 20.00 & 17.17 & 11.67 & 13.49 & 22.71 & 22.03 \\
LongAlpaca & 33.33 & 29.81 & 27.50 & 23.38 & 24.17 & 20.44 & 13.33 & 13.45 & 24.58 & 21.77 \\
LongAlign & 31.67 & 28.06 & 27.50 & 26.22 & 20.83 & 17.31 & 14.17 & 15.02 & 23.54 & 21.65 \\
LongMab-SFT & 30.00 & 29.31 & 30.00 & 28.03 & 14.17 & 13.05 & 10.00 & 10.60 & 21.04 & 20.25 \\
LongReward-PO & 36.67 & \underline{35.04} & 34.17 & 28.77 & \underline{25.83} & 20.85 & 11.67 & 13.91 & 27.09 & 24.64 \\
SeaLong-PO & 31.67 & 30.80 & \textbf{39.17} & \textbf{36.82} & 19.17 & 17.42 & 13.33 & 13.95 & 25.84 & 24.75 \\
Logo-PO & 33.33 & 27.99 & 35.83 & 30.70 & \textbf{28.33} & 15.54 & \underline{15.83} & 12.90 & \underline{28.33} & 21.78 \\
LongFaith-PO & \underline{38.44} & \textbf{35.33} & 35.83 & \underline{32.11} & 21.83 & \underline{21.15} & 14.17 & \textbf{16.11} & 27.57 & \underline{26.18} \\
LongMab & \textbf{40.00} & 34.31 & \underline{37.50} & 32.09 & 23.33 & \textbf{22.65} & \textbf{17.50} & \underline{16.07} & \textbf{29.58} & \textbf{26.28} \\
\hline
\end{tabular}
}
\caption{Performance of different baselines on datasets from LV-Eval across varying context lengths from 16k to 128k. The \textbf{best} and \underline{second best} results are highlighted.}
\label{tab:appendix_lv_eval}
\end{table*}

\begin{figure}[t]
  \centering
  \subfigure[Ground truth recall.]{
    \includegraphics[width=0.46\linewidth]{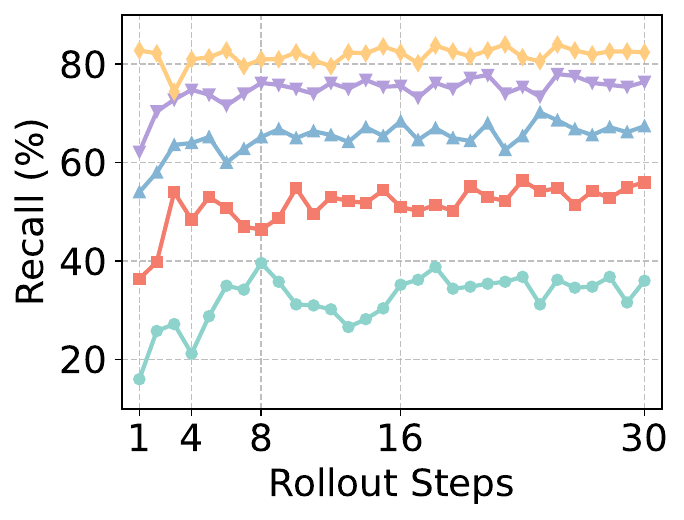}
    \label{fig:topk_recall}
  }
  \subfigure[SubEM of responses.]{
    \includegraphics[width=0.46\linewidth]{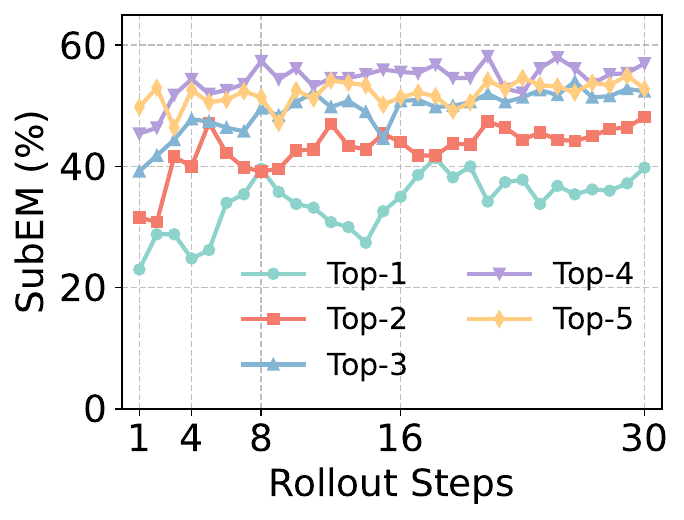}
    \label{fig:topk_subem}
  }
  \caption{Ground truth recall and subem scores with varying numbers of selected chunks ($K$).}
  \label{fig:top-k}
\end{figure}


\begin{table*}[ht]
  \centering
  \setlength{\fboxsep}{0.7pt}
  \resizebox{\linewidth}{!}{
  \begin{tabular}{p{1\textwidth}}
    \hline
    \textbf{Query:} What province shares a border with the province where Lago District is located? \; \\
    \textbf{Ground Truth: Cabo Delgado Province } \\
    \textbf{Document:} \ldots \\
    \textbf{Passage 7: Niassa Reserve} \\
    \ldots \colorbox{lightbluegreen}{Niassa Reserve is a nature reserve in Cabo Delgado Province and Niassa Province}, Mozambique. Covering over 42,000 square kilometres (10,000,000 acres), it is the largest protected area in the country...\\
    \textbf{Passage 12: Lago District} \\
    \ldots \colorbox{lightbluegreen}{Lago District is a district of Niassa Province} in north-western Mozambique. The principal town is Metangula. \colorbox{lightred}{Lago District is bordered to the west by Lake Nyasa}, on the south by Lichinga, on the east by Sanga District.  Lago shares its northern boundary with Tanzania... \\
    \hline
    \multicolumn{1}{c}{\textbf{LongFaith-PO:}} \\
    To determine which province shares a border with the province where Lago District is located, we need to analyze the information provided in the documents. First, we identify the location of Lago District. According to the document, Lago District is located in Niassa Province in northwestern Mozambique [12]. Next, we look for provinces that share a border with Niassa Province. The document states that \colorbox{lightred}{Lago District is bordered to the west by Lake Nyasa}, which implies that it shares a border with the province that includes Lake Nyasa. \colorbox{lightred}{Lake Nyasa is located in Malawi}, which is a neighboring country to Mozambique. Therefore, the province that shares a border with Niassa Province is Malawi. Thus, the answer to the question is Malawi. \colorbox{lightred}{The answer is Malawi}. \\
    \hline
    \multicolumn{1}{c}{\textbf{\method{}:}} \\
    Reasoning: 
    To find the province that shares a border with the province where Lago District is located, we need to identify the location of Lago District first. According to Passage 12, \colorbox{lightbluegreen}{Lago District is a district of Niassa Province} in north-western Mozambique. Next, we need to determine which provinces share a border with Niassa Province. Since Lago District is located in north-western Mozambique, it is likely to share borders with provinces in the neighboring countries or regions. In Passage 7, it is mentioned that \colorbox{lightbluegreen}{Niassa Reserve is located in Cabo Delgado Province and Niassa Province}, Mozambique. This indicates that \colorbox{lightbluegreen}{Niassa Province shares a border with Cabo Delgado Province}. Based on this information, we can conclude that the province that shares a border with the province where Lago District is located is likely to be Cabo Delgado Province.
    \colorbox{lightbluegreen}{Answer: Cabo Delgado Province}.\\
    \hline
  \end{tabular}
  }
  \caption{\label{case_study}
  Case study. We randomly sample an example from the MuSiQue dataset to analyze model behavior. Different colors are used to represent noise information or key evidence. Specifically, \colorbox{lightbluegreen}{green} indicates critical evidence supporting the ground truth, while \colorbox{lightred}{pink} represents distracting information.}
\end{table*}

\subsection{Statistics of Testsets}\label{appendix:testsets}
Table~\ref{tab:benchmark_statistics} shows the details of testsets from LongBench and InfiniteBench.

\subsection{Parameter Sensitivity Analysis}\label{appendix:param_analysis}
In this section, we conduct a comprehensive sensitivity analysis on key hyperparameters used in \method{}, including the number of rollout steps ($T$), the number of selected chunks ($K$), and the exploration–exploitation balance factor ($\alpha$).

\textbf{Rollout Steps Analysis.}
For the rollout step analysis, we set the number of selected chunks to $K=4$ and vary the rollout steps among ${4, 8, 16, 30}$. For each configuration, we construct preference pairs based on the responses collected up to that step. As shown in Table~\ref{tab:rollout_step}, performance consistently improves with an increasing number of rollout steps, suggesting that \method{} progressively refines its sampling process and remains robust throughout iterative rollouts.

\textbf{Selected Chunks Analysis.}
Next, we analyze the sensitivity of \method{} to the number of selected chunks ($K$). We fix the rollout steps at 30 and vary $K$ from 1 to 5, constructing the corresponding preference datasets for each setting. As shown in Table~\ref{tab:top_k_appendix}, performance improves as $K$ increases, reaches its peak at $K=4$, and then slightly declines. To better interpret this pattern, we further track the evolution of ground truth recall and SubEM scores throughout the rollout process (Figure~\ref{fig:topk_recall} and Figure~\ref{fig:topk_subem}). We observe that while selecting more chunks enhances evidence coverage, it also introduces redundant or noisy information, which eventually degrades response quality. Overall, $K=4$ strikes an optimal balance between evidence coverage and noise control, yielding the most effective preference pairs for DPO training.

\textbf{Exploration–Exploitation Balance Factor Analysis.}
Finally, we investigate the effect of the balance coefficient $\alpha$ in the UCB-guided sampling strategy of \method{}. We vary $\alpha$ from 0.2 to 5.0 while keeping other settings fixed. As shown in Table~\ref{tab:alpha}, performance peaks at $\alpha=1.0$ and drops when $\alpha$ is too small or too large. A smaller $\alpha$ biases the model toward exploitation, repeatedly selecting high-reward chunks and yielding preference pairs with limited diversity. Conversely, a larger $\alpha$ emphasizes exploration, increasing the chance of sampling noisy or irrelevant chunks, which degrades the overall quality of responses. Setting $\alpha=1.0$ achieves the best balance between exploration and exploitation, leading to diverse and stable sampling.

\subsection{Generalization Ability of \method{}}
In this section, we further investigate the generalization capability of \method{} beyond the settings reported in the main paper. Our analysis covers two aspects: (1) generalization across model sizes and types, and (2) generalization to longer-context scenarios.

\textbf{Generalization across model sizes and types.}
To assess whether \method{} remains effective for larger and more recent models, we additionally apply it to Qwen2.5-14B-Instruct and Qwen3-8B using exactly the same training hyperparameters. As shown in Table~\ref{tab:qwen2_14b_overall} and Table~\ref{tab:qwen3_overall}, \method{} yields consistent improvements across all datasets and model scales. In particular, \method{} brings an average F1 gain of 6.3\% over the baselines on the larger Qwen2.5-14B-Instruct model, and still delivers a 2.6\% F1 improvement on the latest reasoning model, Qwen3-8B.

\textbf{Generalization to longer contexts.}
Although \method{} is trained on 8k-16k context length, we examine whether the resulting model generalizes to substantially longer context at test time. We evaluated the performance of the resulting model on the HotPotWikiQA-mixup dataset from the LV-Eval benchmark, spanning a wide range of context lengths from 16k up to 128k tokens. As presented in Table~\ref{tab:appendix_lv_eval}, the model fine-tuned with \method{} demonstrates consistent and notable performance gains across all evaluated length regimes. Specifically, it achieves an average F1 gain of 4.2\% over vanilla LLM. These results confirm that the enhanced ability to locate and utilize key information, which is instilled by \method{}, successfully extrapolates to context lengths exceeding 100k tokens.

\subsection{Case Study}\label{appendix:casestudy}
In this section, we present a case from the MuSiQue evaluation set to illustrate how \method{} captures key information from long contexts and alleviates the ``lost-in-the-middle'' problem.

As shown in Table~\ref{case_study}, we present a two-hop question in the MuSiQue dataset. To answer it, the LLM must first identify which province the Lago District belongs to, and then determine which provinces border it based on the context. LongFaith-PO is misled by noisy information near the two ends of the long context and fails to recognize the key evidence appearing in the middle of the long context (specifically, around the 45\% mark of the context), leading to an incorrect answer. In contrast, \method{} enhances the LLM's ability to identify and integrate evidence scattered across different parts of the context, enabling more accurate multi-hop reasoning and ultimately yielding the correct answer.

\end{document}